\documentclass{llncs}
\usepackage[american]{babel}
\usepackage[T1]{fontenc}
\usepackage{times}
\usepackage{graphicx}
\usepackage{booktabs}
\usepackage{wrapfig}
\usepackage{rotating}

\newcommand{\emoji}[1]{{\includegraphics{emoji_images/hires/#1.pdf}}}

\begin{document}

\title{
N-GrAM: New Groningen Author-profiling Model}
\subtitle{Notebook for PAN at CLEF 2017}

\author{Angelo Basile \and Gareth Dwyer \and  Maria Medvedeva \and \\Josine Rawee \and Hessel Haagsma \and Malvina Nissim}
\institute {University of Groningen, Groningen, The Netherlands \\
\email{\{a.basile,g.t.b.dwyer,j.n.rawee\}@student.rug.nl, \\\url{medvmr@gmail.com},\\
 \{hessel.haagsma,m.nissim\}@rug.nl}}

\maketitle

\begin{abstract}

We describe our participation in the PAN 2017 shared task on Author Profiling, identifying authors' gender and language variety for English, Spanish, Arabic and Portuguese. We describe both the final, submitted system, and a series of negative results. Our aim was to create a single model for both gender and language, and for all language varieties. Our best-performing system (on cross-validated results) is a linear support vector machine (SVM) with word unigrams and character 3- to 5-grams as features. A set of additional features, including POS tags, additional datasets, geographic entities, and Twitter handles, hurt, rather than improve, performance. Results from cross-validation indicated high performance overall and results on the test set confirmed them, at 0.86 averaged accuracy, with performance on sub-tasks ranging from 0.68 to 0.98. 
\end{abstract}

\section{Introduction}

With the rise of social media, more and more people acquire some kind of on-line presence or persona, mostly made up of images and text. This means that these people can be considered authors, and thus that we can profile them as such. Profiling authors, that is, inferring personal characteristics from text, can reveal many things, such as their age, gender, personality traits, location, even though writers might not consciously choose to put indicators of those characteristics in the text. The uses for this are obvious, for cases like targeted advertising and other use cases, such as security, but it is also interesting from a linguistic standpoint. 

In the shared task on author profiling \cite{rangel:2017}, organised within the PAN framework \cite{potthast:2017a}, the aim is to infer Twitter users' gender and language variety from their tweets in four different languages: English, Spanish, Arabic, and Portuguese. Gender consists of a binary classification (male/female), whereas language variety differs per language, from 2 varieties for Portuguese (Brazilian and Portugal) to 7 varieties for Spanish (Argentina, Chile, Colombia, Mexico, Peru, Spain, Venezuela). The challenge is thus to classify users along two very different axes, and in four highly different languages -- forcing participants to either build models that can capture these traits very generally (language-independent) or tailor-make models for each language or subtask.

Even when looking at the two tasks separately, it looks like the very same features could be reliable clues for classification. Indeed, for both profiling authors on Twitter as well as for discriminating between similar languages, word and character n-grams have proved to be the strongest predictors of gender as well as language varieties. For language varieties discrimination, the systems that performed best at the DSL shared tasks in 2016 (on test set B, i.e. social media) used word/character n-grams, independently of the algorithm \cite{malmasi-EtAl:2016:VarDial3}. The crucial contribution of these features was also observed by \cite{medvedeva2017en,bestgen2017improving}, who participated in the 2017~DSL~shared task with the two best performing systems. For author profiling, it has been shown that tf-idf weighted n-gram features, both in terms of characters and words, are very successful in capturing especially gender distinctions \cite{overview:2016}. If different aspects such as language variety and gender of a speaker on Twitter might be captured by the same features, can we build a single model that will characterise both aspects at once?

In the context of the PAN~2017 competition on user profiling we therefore experimented with enriching a basic character and word n-gram model by including a variety of features that we believed should work. We also tried to view the task jointly and model the two problems as one single label, but single modelling worked best.

In this paper we report how our final submitted system works, and provide some general data analysis, but we also devote substantial space to describing what we tried (under which motivations), as we believe this is very informative towards future developments of author profiling systems.

\section{Final System}
\label{sec:system}
After an extensive grid-search we submitted as our final run, a simple SVM system (using the scikit-learn LinearSVM implementation) that uses character 3- to 5-grams and word 1- to 2-grams with tf-idf weighting with sublinear term frequency scaling, where instead of the standard term frequency the following is used: 

\begin{center}
$1 + \log(tf)$
\end{center}

\noindent We ran the grid search over both tasks and all languages on a 64-core machine with 1 TB RAM (see Table~\ref{tab:gridsearch} for the list of values over which the grid search was performed). The full search took about a day to complete. In particular, using min\_df=2 (i.e. excluding all terms that are used by only one author) seems to have a strong positive effect and greatly reduces the feature size as there are many words that appear only once. The different optimal parameters for different languages provided only a slight performance boost for each language. We decided that this increase was too small to be significant, so we decided to use a single parameter set for all languages and both tasks.

\begin{table}
\begin{center}
\caption{Results (accuracy) for the 5-fold cross-validation\label{tab:finalresults}}
\begin{tabular}{lrrr}
\toprule
\textbf{Language} & \textbf{Variety} & \textbf{Gender} & \textbf{Both} \\
\midrule
Arabic & 0.831 & 0.800 & 0.683\\
English & 0.898 & 0.823 & 0.742\\
Spanish & 0.962 & 0.832 & 0.803\\
Portuguese & 0.981 & 0.845 & 0.828\\
\bottomrule
\end{tabular}
\end{center}
\end{table}

\begin{table}[hbtp]
\centering
\caption{A list of values over which we performed the grid search.\label{tab:gridsearch}
}
\begin{tabular}{lll}
\toprule
\textbf{Name} & \textbf{Values}     & \textbf{Description}                                \\ \midrule
lowercase     & True, False         & Lowercase all words                                 \\
max\_df       & 0.01, None          & Exclude terms that appear in more than $n$ documents  \\
min\_df       & 1, 2, 3             & Exclude terms that appear in fewer than $n$ documents \\
use\_idf      & True, False         & Use Inverse Document Frequency weighting            \\
sublinear\_tf & True, False         & Replace term frequency (tf) with $1 + log(tf)$        \\
C             & 0.1, 0.5, 1, 1.5, 5 & Penalty parameter for the SVM           \\           
\bottomrule
\end{tabular}
\end{table}

\section{Data Analysis}

The training dataset provided consist of 11400 sets of tweets, each set representing a single author. The target labels are evenly distributed across variety and gender. The labels for the gender classification task are `male' and `female'. Table \ref{tab:langvariety} shows the labels for the language variation task and also shows the data distribution across languages.

\begin{figure}
 \centering
 \includegraphics[scale=0.2]{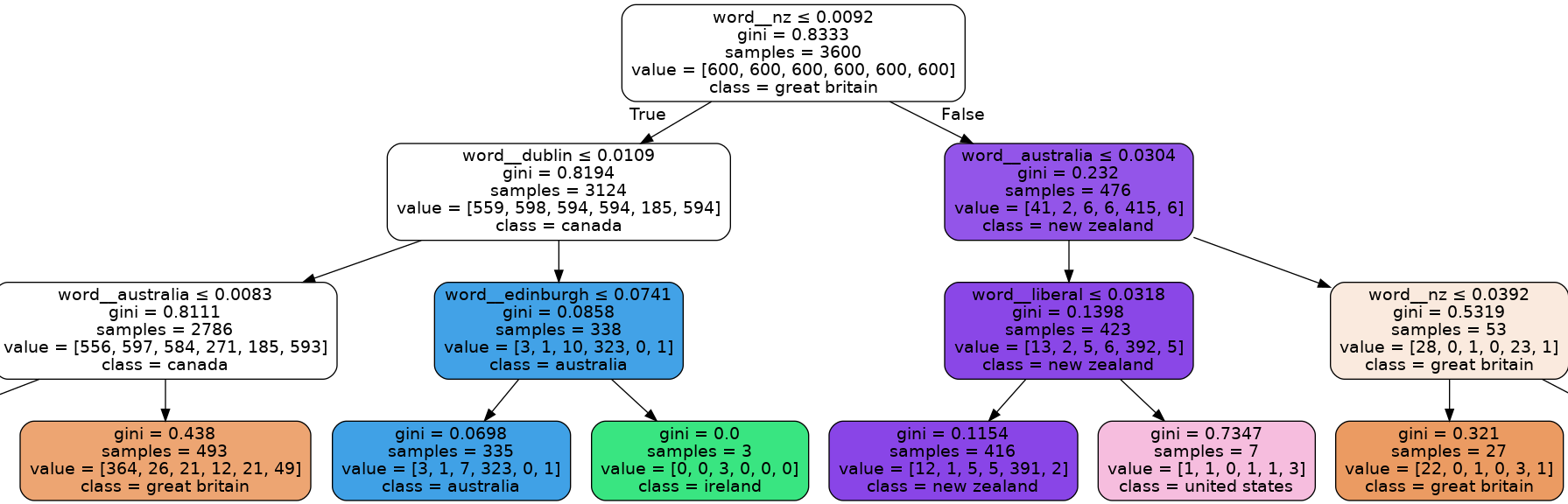}
 \caption{Decision Tree output}
 \label{fig:dtree1}
\end{figure}

We produced two visualisations, one per label (i.e. variety and gender), in order to gain some insights that could help the feature engineering process. For the variety label we trained a decision tree classifier using word unigrams: although the performance is poor (accuracy score of 0.63) this setup has the benefit of being easy to interpret: Figure \ref{fig:dtree1} shows which features are used for the first splits of the tree.

We also created a visualisation of the English dataset using the tool described in \cite{kessler2017scattertext}, and comparing the most frequent words used by males to those used by females. The visualisation shown in Figure \ref{fig:scatter} indicates several interesting things about the gendered use of language. The words used often by males and very seldom by females are often sport-related, and include words such as ``league'', and ``chelsea''. There are several emojis that are used frequently by females and infrequently by males, e.g.  ``\emoji{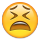}'', ``\emoji{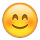}'', 
as well as words like ``kitten'', ``mom'', ``sister'' and ``chocolate''. In the top right of the visualisation we see words like ``trump'' and ``sleep'', which indicates that these words are used very frequently, but equally so by both genders. This also shows that distinguishing words include both time-specific ones, like ``gilmore'' and ``imacelebrityau'', and general words from everyday life, which are less likely to be subject to time-specific trends, like ``player'', and ``chocolate''.

\begin{table}[hbtp]
\caption{\label{tab:langvariety}Variety labels per language}
\begin{center}
\begin{tabular}{lp{7cm}}
\toprule
\textbf{Language} & \textbf{Variety} \\
\midrule
English & Australia, Canada, Great Britain, Ireland, New Zealand, United States \\
Spanish & Argentina, Chile, Colombia, Mexico, Peru, Spain, Venezuela \\
Portuguese & Brazil, Portugal \\
Arabic & Egypt, Gulf, Levant, Maghreb \\
\bottomrule
\end{tabular}
\end{center}
\end{table}

\section{Alternative Features and Methods: An Analysis of Negative Results}

This section is meant to highlight all of the potential contributions to the systems which turned out to be detrimental to performance, when compared to the simpler system that we have described in Section~\ref{sec:system}.  We divide our attempts according to the different ways we attempted to enhance performance: manipulating the data itself (adding more, and changing preprocessing), using a large variety of features, and changing strategies in modelling the problem by using different algorithms and paradigms. All reported results are on the PAN 2017 training data using five-fold cross-validation, unless otherwise specified.

\subsection{Supplementary Data and Features}

\subsubsection{Adding Previous PAN Data}
We extended the training dataset by adding data and gender labels from the PAN 16 Author Profiling shared task \cite{overview:2016}. However, the additional data consistently resulted in lower cross-validation scores than when using only the training data provided with the PAN 17 task. One possible explanation for this is that our unigram model captures aspects that are tied specifically to the PAN 17 dataset, because it contains topics that may not be present in datasets that were collected in a different time period.
To confirm this, we attempted to train on English data from PAN 17 and predict gender labels for the English data from PAN 16, as well as vice versa. Training on the PAN 16 data resulted in an accuracy score of 0.754 for the PAN 17 task, and training on PAN 17 gave an accuracy score of 0.70 for PAN 16, both scores significantly lower than cross-validated results on data from a single year.

\subsubsection{Using the Twitter 14k dataset}
We attempted to classify the English tweets by Gender using only the data collected by \cite{bamman2014gender}. This dataset consists of aggregated word counts by gender for about 14,000 Twitter users and 9 million Tweets. We used this data to calculate whether each word in our dataset was a `male' word (used more by males), or a `female' word, and classified users as male or female based on a majority count of the words they used. Using this method we achieved 71.2 percent accuracy for the English gender data, showing that this simple method can provide a reasonable baseline to the gender task.

\subsubsection{Tokenization}

We experimented with different tokenization techniques for different languages, but our average results did not improve, so we decided to use the default scikit-learn tokenizer.

\subsubsection{POS tags}

We tried adding POS-tags to the English tweets using the \textit{spaCy}\footnote{https://spacy.io/} tagger: compared to the model using unigrams only the performances dropped slightly for gender and a bit more for variety:

\begin{table}[hbtp]
\caption{Results (accuracy) on the English data for Gender and Variety with and without part of speech tags.}
\centering
\begin{center}
\begin{tabular}{lrr}
\toprule
 & \textbf{Gender} & \textbf{Variety}\\
\hline
Unigrams & 0.826 & 0.895\\
Unigrams + Part-of-Speech & 0.818 & 0.853\\
\bottomrule
\end{tabular}
\end{center}
\end{table}

It is not clear whether the missed increase in performance is due to the fact that the data are not normal (i.e. the tokenizer is not Twitter specific) or to the fact that POS tags confuse the classifier. Considering the results we decided not to include a POS-tagger in the final system.

\subsubsection{Emojis}
(\emoji{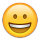}\emoji{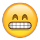}\emoji{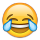}\emoji{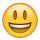}\emoji{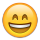}\emoji{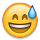}\emoji{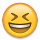}\emoji{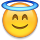}\emoji{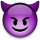}\emoji{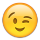}\emoji{1F60A}\emoji{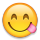})

In April 2015, SwiftKey did an extensive report\footnote{https://blog.swiftkey.com/americans-love-skulls-brazilians-love-cats-swiftkey-emoji-meanings-report/} on emoji use by country. They discovered that emoji use varies across languages and across language varieties. For example, they found that Australians use double the average amount of alcohol-themed emoji and use more junk food and holiday emoji than anywhere else in the world.

We tried to leverage these findings but the results were disappointing. We used a list of emojis\footnote{http://www.unicode.org/emoji/charts/full-emoji-list.html} as a vocabulary for the td/idf vectorizer. Encouraged by the results of the SwiftKey report, we tried first to use emojis as the only vocabulary and although the results are above the baseline and also quite high considering the type of features, they were still below the simple unigram model. Adding emojis as extra features to the unigram model also did not provide any improvement. 

Since emojis are used across languages we built a single model for the four languages. We trained the model for the gender label on English, Portuguese and Arabic and tested it on Spanish: the system scored 0.67 in accuracy.

\subsubsection{Excluding Specific Word Patterns}
We looked at accuracy scores for the English gender and variety data more closely. We tried different representations of the tweet texts, to see what kind of words were most predictive of variety and gender. Specifically, we look at using only words that start with an uppercase letter, only words that start with a lowercase letter, only Twitter handles (words that start with an "@") and all the text excluding the handles. 

It is interesting that the accuracies are so high although we are using only a basic unigram model, without looking at the character n-grams that we include in our final model. Representing each text only by the Twitter handles used in that text results in 0.77 accuracy for variety, probably because users tend to interact with other users who are in the same geographic area. However, excluding handles from the texts barely decreases performance for the variety task, showing that while the handles can be discriminative, they are not necessary for this task. It is also interesting to note that for this dataset, looking only at words beginning with an uppercase character results in nearly the same score for the Gender task as we get when using all of the available text, while using only lowercase words decreases performance. The opposite is true for the variety task, where using lowercase-only words results in as good performance as using all the text, but using only uppercase words decreases accuracy by over 10 percent.

\begin{table}[hbtp]
\centering
\caption{Results (accuracy) on the English data for Gender and Variety when excluding certain words. We preprocessed the text to exclude the specified word-patterns and then vectorized the resulting text with tf-idf. Classification was done using an SVM with a linear kernel over five-fold cross-validation.}
\label{my-label}
\begin{tabular}{lll}
\toprule
                & \textbf{Gender} & \textbf{Variety} \\
\midrule
All text        & 0.816           & 0.876            \\
Handles only    & 0.661           & 0.769            \\
Exclude handles & 0.814           & 0.869            \\
Uppercase only  & 0.802           & 0.767            \\
Lowercase only  & 0.776           & 0.877            \\
\bottomrule
\end{tabular}
\end{table}

\subsubsection{Place Names and Twitter Handles}
We tried using the counts of geographical names related to the language varieties were as a feature. We also treated this list of locations as vocabulary for our model. Both these approaches did not improve our model. 

We then tried enriching the data to improve the Unigram model. For each of the language varieties, we obtained 100 geographical location names, representing the cities with the most inhabitants. When this location was mentioned in the tweet, the language variety the location was part of was added to the tweet.

We attempted to use Twitter handles in a similar manner. The 100 most-followed Twitter users per language variety were found and the language variety was added to the text when one of its popular Twitter users was mentioned.

Unfortunately, this method did not improve our model. We suspect that the information is being captured by the n-gram model, which could explain why this did not improve performance. 

\subsubsection{GronUP Combos}

We have tried the partial setup of last year's winning system, GronUP \cite{gronup}, with the distinction that we had to classify language variety instead of age groups. We have excluded the features that are language-dependent (i.e.\ pos-tagging and misspelling/typos), and experimented with various feature combinations of the rest while keeping word and character \textit{n}-grams the same. We achieved average accuracy from 0.810 to 0.830, which is clearly lower than our simple final model.

\subsection{Modelling}

\subsubsection{Joint Prediction}
\setlength{\tabcolsep}{4pt}
 \begin{wraptable}{r}{3.4cm}
\vspace*{-1.1cm}
 \centering
\centering
\caption{\label{tab:jointpred}Results (accuracy) for the joint prediction of Gender and Variety.}
\begin{tabular}{lc}
\toprule
& \textbf{Joint} \\
\midrule
Arabic      &  0.630 \\
English     &  0.645 \\
Spanish     &  0.686 \\
Portuguese  &  0.792 \\
\bottomrule
\end{tabular}
\vspace*{-1cm}
\end{wraptable}
We tried to build a single model that predicts at the same time both the language variety and the gender of each user: as expected (since the task is harder) the performance goes down when compared to a model trained independently on each label. However, as highlighted in Table \ref{tab:jointpred}, the results are still surprisingly high. To train the system we simply merged the two labels.

\subsubsection{Different Approaches}
We experimented with Facebook's FastText system, which is an out-of-the-box supervised learning classifier \cite{joulin2016bag}. We used only the data for the English gender task, trying both tweet-level and author-level classification. We pre-processed all text with the NLTK Tweet Tokenizer and used the classification-example script provided with the FastText code base. Training on 3,000 authors and testing on 600 authors gave an accuracy score of 0.64. Changing the FastText parameters such as number of epochs, word n-grams, and learning rate showed no improvement. We achieved an accuracy on 0.79 when we attempted to classify on a per-tweet basis (300,000 tweets for training and 85,071 for test), but this is an easier task as some authors are split over the training and test sets. There are various ways to summarise per-tweet predictions into author-predictions, but we did not experiment further as it seemed that the SVM system worked better for the amount of data we have. 

In the final system we used the SVM classifier because it outperformed all the others that we tried. Table \ref{tab:otherclfs} highlights the results.

\begin{table}[hbtp]
\centering
\caption{\label{tab:otherclfs} Performances per classifier: DT: Decision Tree; MLP: Multi-Layer Perceptron, NB: Naive Bayes.}
\begin{tabular}{llllllll}
\toprule
     & \multicolumn{3}{c}{\textbf{Gender}} & & \multicolumn{3}{c}{\textbf{Variety}} \\
\toprule
 & DT     & MLP     & NB  &   & DT      & MLP     & NB     \\ \midrule
Arabic   &  0.619       &   0.619    & 0.699   & ~~    &  0.685        & 0.813       & 0.729\\
English   &  0.635       &   0.798    & 0.745    &~~    &  0.689        & 0.845       & 0.696\\
Spanish   &  0.608       &   0.783    & 0.677    &~~    &  0.782        & 0.944       & 0.829\\
Portuguese   &  0.714       &   0.813    & 0.676    &~~    &  0.955        & 0.986       & 0.983\\
\bottomrule
\end{tabular}
\end{table}

\section{Results on Test Data}
For the final evaluation we submitted our system, N-GrAM, as described in Section~2. Overall, N-GrAM came first in the shared task, with a score of 0.8253 for gender 0.9184 for variety, a joint score of 0.8361 and an average score of 0.8599 (final rankings were taken from this average score \cite{rangel:2017}). For the global scores, all languages are combined. We present finer-grained scores showing the breakdown per language in Table~\ref{tab:testresults}. We compare our gender and variety accuracies against the LDR-baseline \cite{rangel2017low}, a low dimensionality representation especially tailored to language variety identification, provided by the organisers. The final column, \textit{+ 2nd} shows the difference between N-GrAM and that achieved by the second-highest ranked system (excluding the baseline).

Results are broken down per language, and are summarised as both \textit{joint} and \textit{average} scores. The joint score is is the percentage of texts for which both gender and variety were predicted correctly at the same time. The average is calculated as the mean over all languages.

N-GrAM ranked first in all cases except for the language variety task. In this case, the baseline was the top-ranked system, and ours was second by a small margin. Our system significantly out-performed the baseline on the joint task, as the baseline scored significantly lower for the gender task than for the variety task.

\setlength{\tabcolsep}{4pt}
\begin{table}
\caption{\label{tab:testresults} Results (accuracy) on the test set for variety, gender and their joint prediction.}
\begin{center}
\begin{tabular}{llrrrrrr}
\toprule
\bf Task & \bf System & \bf Arabic & \bf English & \bf Portuguese & \bf Spanish & \bf Average & + 2nd \\
\hline
Variety & N-GrAM & \textbf{0.8313} & 0.8988 & 0.9813 & 0.9621 & 0.9184 & 0.0013\\
 & LDR & 0.8250 & \textbf{0.8996} & \textbf{0.9875} & \textbf{0.9625} & \textbf{0.9187} & \\
\hline
Gender & N-GrAM & \textbf{0.8006} & \textbf{0.8233} & \textbf{0.8450} & \textbf{0.8321} & \textbf{0.8253} & 0.0029\\
 & LDR & 0.7044 & 0.7220 & 0.7863 & 0.7171 & 0.7325 & \\
\hline
Joint & N-GrAM & \textbf{0.6831} & \textbf{0.7429} & \textbf{0.8288} & \textbf{0.8036} & \textbf{0.7646} & 0.0101\\
 & LDR & 0.5888 & 0.6357 & 0.7763 & 0.6943 & 0.6738 & \\
 \bottomrule
\end{tabular}
\end{center}
\end{table}

\section{Conclusion}
We conclude that, for the current author profiling task, a seemingly simple system using word and character n-grams and an SVM classifier proves very hard to beat. Indeed, N-GrAM turned out to be the best-performing out of the 22 systems submitted in this shared task. Using additional training data, `smart' features, and hand-crafted resources hurts rather than helps performance. A possible lesson to take from this would be that manually crafting features serves only to hinder a machine learning algorithm's ability to find patterns in a dataset, and perhaps it is better to focus one's efforts on parameter optimisation instead of feature engineering.

However, we believe that this is too strong a conclusion to draw from this limited study, since several factors specific to this setting need to be taken into account. For one, a support vector machine clearly outperforms other classifiers, but this does not mean that this is an inherently more powerful. Rather, we expect that an SVM is the best choice for the given amount of training data, but with more training data, a neural network-based approach would achieve better results. 

\begin{sidewaysfigure}
 \centering
 \includegraphics[scale=0.5]{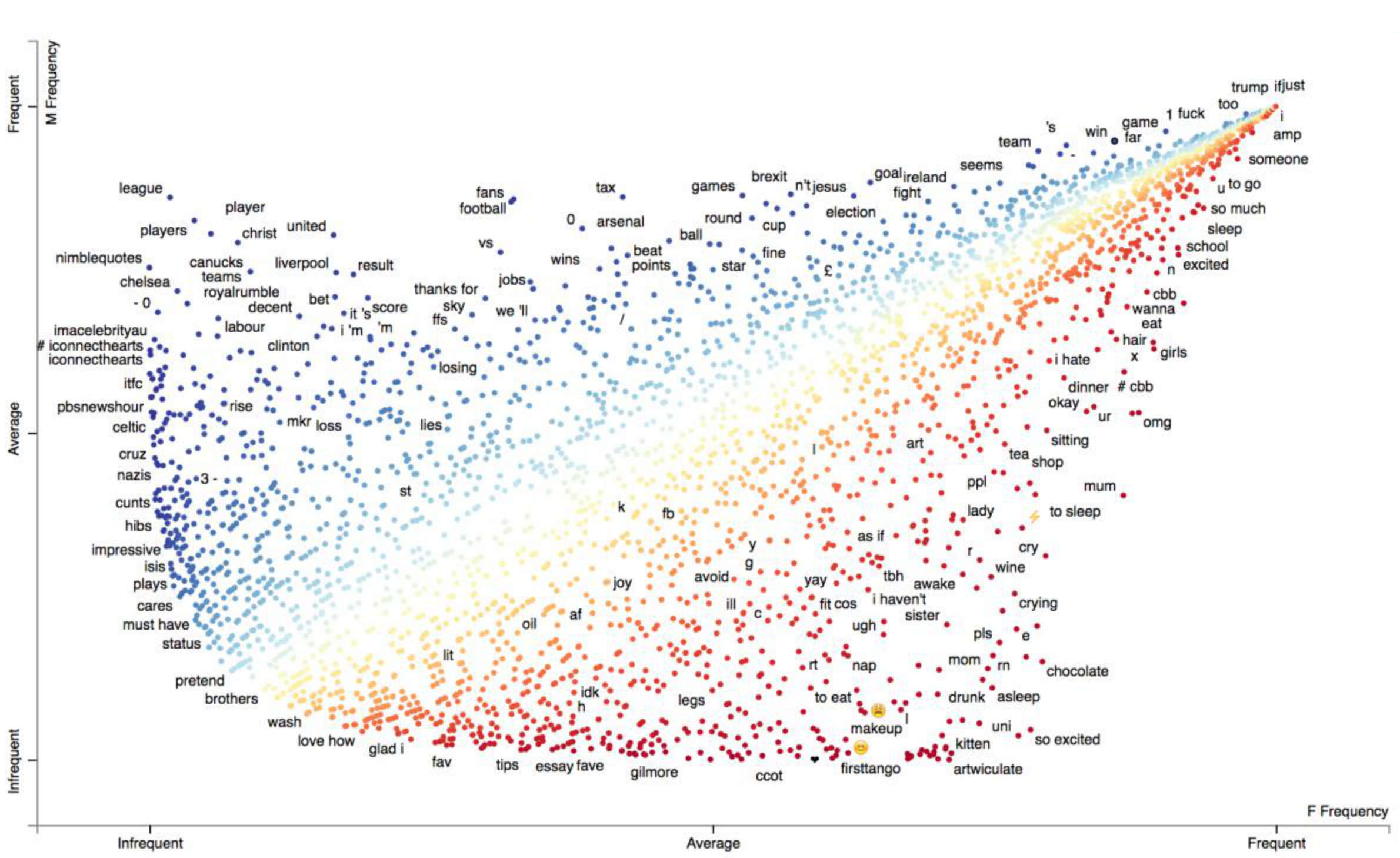}
 \caption{Scatter plot of terms commonly used by male and female English speakers.}
 \label{fig:scatter}
\end{sidewaysfigure}

Regarding the frustrating lack of benefit from more advanced features than n-grams, a possible explanation comes from a closer inspection of the data. Both the decision tree model (see Figure~\ref{fig:dtree1}) and the data visualisation (see Figure~\ref{fig:scatter}) give us an insight in the most discriminating features in the dataset. In the case of language variety, we see that place names can be informative features, and could therefore be used as a proxy for geographical location, which in turn serves as a proxy for language variety. Adding place names explicitly to our model did not yield performance improvements, which we take to indicate that this information is already captured by n-gram features. Whether and how geographical information in the text can be useful in identifying language variety, is a matter for future research.

In the case of gender, many useful features are ones that are highly specific to the Twitter platform (\textit{\#iconnecthearts}), time (\textit{cruz}), and topics (\textit{pbsnewshour}) in this dataset, which we suspect would not carry over well to other datasets, but provide high accuracy in this case. Conversely, features designed to capture gender in a more general sense do not yield any benefit over the more specific features, although they would likely be useful for a robust, cross-dataset system. These hypotheses could be assessed in the future by testing author profiling systems in a cross-platform, cross-time setting.

\bibliographystyle{splncs03}
\begin{raggedright}
\bibliography{bibliography}
\end{raggedright}

\end{document}